\documentclass[letterpaper, 10 pt, conference]{ieeeconf}  % Comment this line out if you need a4paper

\usepackage{amsmath}
\usepackage{graphicx}
\usepackage{float}
\usepackage{booktabs}
\usepackage{subcaption}
\usepackage{cite}
\IEEEoverridecommandlockouts                              % This command is only needed if 
\title{\LARGE \bf
A Novel Modular Cable-Driven Soft Robotic Arm with Multi-Segment Reconfigurability}

\author{
Moeen Ul Islam$^{1}$, Cheng Ouyang$^{1}$, Xinda Qi$^{2}$,  Azlan Zahid$^{3}$, Xiaobo Tan$^{2}$, and Dong Chen$^{1}$%
\thanks{$^{1}$Moeen Ul Islam, Cheng Ouyang, and Dong Chen are with the Department of Agricultural and Biological Engineering, Mississippi State University, MS, USA.
Emails: {\tt\small \{mu136, co603, dc2528\}@msstate.edu}}%
\thanks{$^{2}$Xinda Qi and Xiaobo Tan are with the Department of Electrical and Computer Engineering, Michigan State University, MI, USA.
Emails: {\tt\small qixinda@msu.edu, xbtan@egr.msu.edu}}%
\thanks{$^{3}$Azlan Zahid is with the  
Texas A\&M AgriLife Research and Extension Center in Dallas, Texas A\&M University, TX, USA.
Email: {\tt\small azlan.zahid@ag.tamu.edu}}%
}

\begin{document}

\maketitle
\thispagestyle{empty}
\pagestyle{empty}

%%%%%%%%%%%%%%%%%%%%%%%%%%%%%%%%%%%%%%%%%%%%%%%%%%%%%%%%%%%%%%%%%%%%%%%%%%%%%%%%
\begin{abstract}
This paper presents a novel, modular, cable-driven soft robotic arm featuring multi-segment reconfigurability. The proposed architecture enables a stackable system with independent segment control, allowing scalable adaptation to diverse structural and application requirements.
The system is fabricated from soft silicone material and incorporates embedded tendon-routing channels with a protective dual-helical tendon structure. Experimental results showed that modular stacking substantially expanded the reachable workspace: relative to the single-segment arm, the three-segment configuration achieved up to a 13-fold increase in planar workspace area and a 38.9-fold increase in workspace volume. Furthermore, this study investigated the effect of silicone stiffness on actuator performance. The results revealed a clear trade-off between compliance and stiffness: softer silicone improved bending flexibility, while stiffer silicone improved structural rigidity and load-bearing stability. These results highlight the potential of stiffness tuning to balance compliance and strength for configuring scalable, reconfigurable soft robotic arms.

\end{abstract}

%%%%%%%%%%%%%%%%%%%%%%%%%%%%%%%%%%%%%%%%%%%%%%%%%%%%%%%%%%%%%%%%%%%%%%%%%%%%%%%%
\section{INTRODUCTION}
Rigid robots are typically made up of stiff structural materials and actuated by electrical actuators or pressurized fluids. Although these systems are technologically mature, they remain limited in safety during human–robot collaboration. In addition, their higher weight, restricted adaptability to dynamic environments, and rigid structure constrain their deployment in delicate or unstructured settings \cite{Whitesides2018-hz}. In contrast, soft robots are fabricated from compliant materials shaped by molding or additive manufacturing methods, enabling improved adaptability in unstructured environments \cite{Lee2017-vb}. Owing to their superior flexibility and inherent compliance compared to rigid robots, research on soft robots has grown rapidly recently \cite{Yasa2023, Zongxing2014}. Unlike traditional rigid manipulators with limited workspace and dexterity, soft robots are well suitable for operation in complex environments and for safe human–robot interaction \cite{Das2019}.

%Unlike traditional rigid manipulators, which are limited in workspace and dexterity, soft robotic arms offer greater flexibility and adaptability, making them particularly suitable for sensitive applications where safe human–robot interaction is essential \cite{Das2019}.

Based on the actuation method, soft robots can be classified as cable-driven, pneumatic, electro-adhesion, electro-active polymer, and shape memory alloy (SMA) actuated systems \cite{Lee2017-vb}. Recent studies have explored diverse implementations, such as SMA coil–based soft robotic arms \cite{Yang2019}, pneumatically extensible soft robotic arms \cite{Chen2019}, and topology-optimized cable-driven grippers \cite{ouyang2025direct, Wang2020, Chen2018}. Among these approaches, cable-driven soft robots use tendons connected to servo or stepper motors, offering lower system weight, quiet operation, and faster response, while maintaining higher power output and payload capacity \cite{Zaidi2021}. Furthermore, cable-driven systems enable precise and independent tendon control through compact motor modules, facilitating accurate motion generation and repeatable deformation \cite{Whitesides2018-hz}.

% In recent years, different kinds of soft robotic manipulators have been developed for specific applications \cite{Chen2021}. These include SMA coil–based soft robotic arms \cite{Yang2019}, pneumatically extensible soft robotic arms \cite{Chen2019}, and topology-optimized cable-driven grippers \cite{ouyang2025direct, Wang2020, Chen2018}. However, these systems are not designed for general-purpose use; rather, they are intended for specific applications due to the fabrication-based limitations of soft robots. Such robots often have a limited workspace and are typically built to serve a single function.

Despite these advances, most existing soft robotic manipulators lack flexibility in both design and configuration \cite{Gilpin2010}. They are typically fabricated as single, continuous structures optimized for a specific task, which limits their scalability and reusability \cite{Elango2015, Lin2023, Gunderman2022}. Once manufactured, such systems provide little adaptability in length, geometry, or functionality, making it difficult to tailor a single system for multiple applications \cite{Knospler2024}. To address these issues, researchers have explored \textit{modular} soft robotic systems that enable enhanced adaptability and reconfiguration \cite{Knospler2024}. In parallel, several studies have investigated configuration-space and workspace modeling of modular soft arms, offering valuable insights into motion planning and kinematic mapping \cite{Chen2025}. However, many existing modular designs rely on suction or magnetic-based mechanisms for attachment and detachment. These approaches often suffer from weak inter-module coupling and limited mechanical reinforcement, leading to reduced payload capacity and compromised structural stability \cite{Zhang2022}. For broader and more versatile applications, modular soft robotic systems should enable simple attachment and detachment of modules, allowing flexible adjustment of arm length, configuration, and even the number of arms according to task-specific requirements \cite{Gilpin2010}.

% To overcome these limitations, modular soft robots have been introduced. These are generally classified into two categories:

% \begin{enumerate}
%     \item \textbf{Fabricated-assembled robots}, where the arrangement and connection of modular units are fixed.
%     \item \textbf{Reconfigurable soft robots}, where modules can be rearranged or replaced based on need \cite{Zhang2020}.
% \end{enumerate}

% Recent research has explored various modular soft robotic systems with enhanced adaptability and stiffness modulation to achieve versatile task capabilities \cite{Knospler2024}. Similarly, studies have investigated configuration-space and workspace modeling of modular soft robotic arms, providing important insights into motion planning and kinematic mapping for such systems \cite{Chen2025}. However, most existing modular designs still rely on suction or magnetic forces for attachment and detachment. While these mechanisms enable locomotion, they generally limit payload capacity \cite{Zhang2022}.

Material composition and density play a key role in determining the strength and durability of soft actuators, particularly during push–pull interactions or object manipulation \cite{Elango2015}. While prior studies have explored various strategies to improve load capacity through structural or architectural modifications \cite{Thanigaivel2025,Li2021}, a systematic, quantitative understanding of how material stiffness alone influences payload performance in modular, cable-driven soft arms remains largely unexplored. 

%In such systems, each module must be capable of independently supporting its payload, as payload capacity is a critical factor influencing real-world manipulation performance \cite{Garbulinski2024}. 

%Prior studies have shown that rigid, vertebrate-inspired architectures enhance stability while retaining flexibility \cite{Thanigaivel2025}, and that structural reinforcement can further improve payload capacity \cite{Li2021}. 
%

%Motivated by these findings, 

This paper proposes a fully \textit{modular}, \textit{cable-driven} soft robotic arm in which each segment can be independently attached or detached together with its corresponding actuation module. Building on this reconfigurable platform, we systematically investigate how silicone stiffness influences actuation behavior and load-bearing capacity, validated through quantitative motion-capture experiments across single- and multi-segment configurations.

\section{HARDWARE DESIGN}
\subsection{Design and Fabrication of Soft Robot}
\begin{figure}[!ht]
    \centering
    \includegraphics[width=0.9\linewidth]{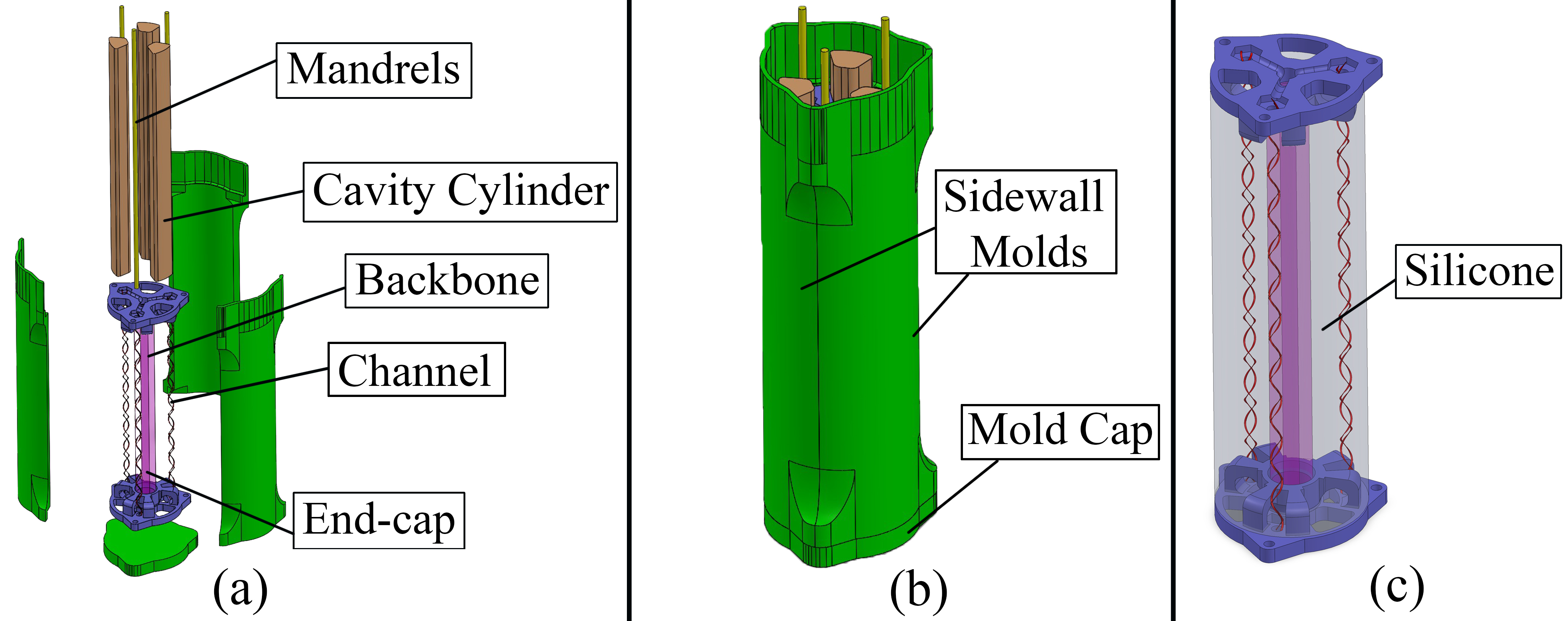}
    \caption{Fabrication process of the silicone-based, cable-driven soft robotic arm: (a) internal assembly components, (b) silicone molding process, and (c) final cured segment.}
    \label{fig:fabrication}
\end{figure}

As shown in Fig.~\ref{fig:fabrication}, each soft robotic arm consists of a flexible internal backbone, cable-routing channels for tendon guidance, and two end caps, forming a cast-silicone cable-driven structure composed of a small number of repeatable mechanical elements \cite{Qi2024}. 

The soft arm fabrication process begins with 3D printing the end caps in PETG using an FDM 3D printer. The end caps were designed with interlocking mechanical features that embed into the silicone during curing, preventing detachment during actuation. They were also shaped to allow modular stacking, enabling additional segments to be mounted on existing ones to extend the arm length. Clear McMaster-Carr Soft-Masterkleer PVC tubing with a 3 mm inner diameter and 6 mm outer diameter was used as the central backbone of the soft robot. The tubing runs through the center of each segment, providing a flexible yet constrained axis for tendon routing. Kevlar threads were wound in a dual helical pattern around a 2 mm metal rod, which served as a mandrel during silicone casting. This process formed channels that act as both tendon guides and local reinforcement within the silicone body. For reusable, attachable, and detachable molds, three 3D-printed sidewalls and a bottom mold cap were created. To reduce weight, increase flexibility, and introduce a hollow cavity in the final soft robot, three 3D-printed cavity cylinders were incorporated into the mold as temporary mandrels. An exploded view of the mold casting components is shown in Fig.~\ref{fig:fabrication}(a). 

%\vspace{-7pt}
% \vspace{-4pt}

To prepare the mold for silicone casting, the PVC tube was cut to the required length, and end caps were attached to both ends of the tube. The 2 mm metal rods, acting as mandrels, were then positioned in place, and the dual helical tendon structure was formed around the mandrels to create the channels for actuation tendons. The two ends of each tendon were knotted and secured to the corresponding end caps. The cavity mandrels were also inserted between the two end caps through their designated holes. This assembled internal structure was placed on top of the bottom mold cap, after which three sidewall mold components were clamped around it, enclosing all internal elements, as shown in Fig.~\ref{fig:fabrication}(b). Liquid silicone was then poured into the enclosed mold from the top. The top end cap contained dedicated inlet holes to allow the silicone to fully fill the volume between the two end caps. After the silicone cured, the mold components were disassembled, and all mandrels were carefully extracted vertically through the top end cap. The final fabricated silicone soft robot is shown in Fig.~\ref{fig:fabrication}(c). To actuate the soft robot, tendons were routed through the internal channels and securely attached to the bottom end cap at designated mounting points, while the opposite ends of the tendons were connected to stepper motors via a pulley mechanism.

\begin{figure*}[!ht]
    \centering
    % Top single figure
    \begin{subfigure}{0.9\textwidth}
        \centering
        \includegraphics[width=\textwidth]{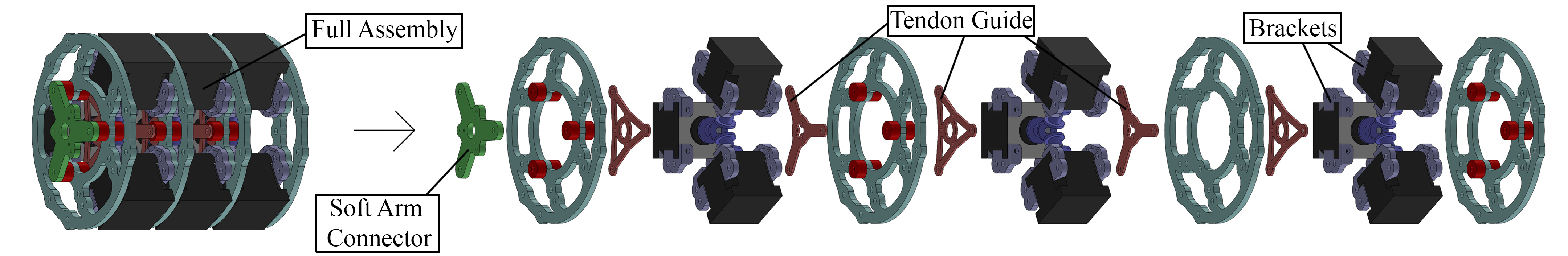}
        \caption{Exploded and assembled views of the actuation module with the motor frame, brackets, and three-segment layout.}
        \label{fig:actuation_module_a}
    \end{subfigure}

    \vspace{0.2em}

    % Bottom three figures
    \begin{subfigure}{0.32\textwidth}
        \centering
        \includegraphics[width=\textwidth]{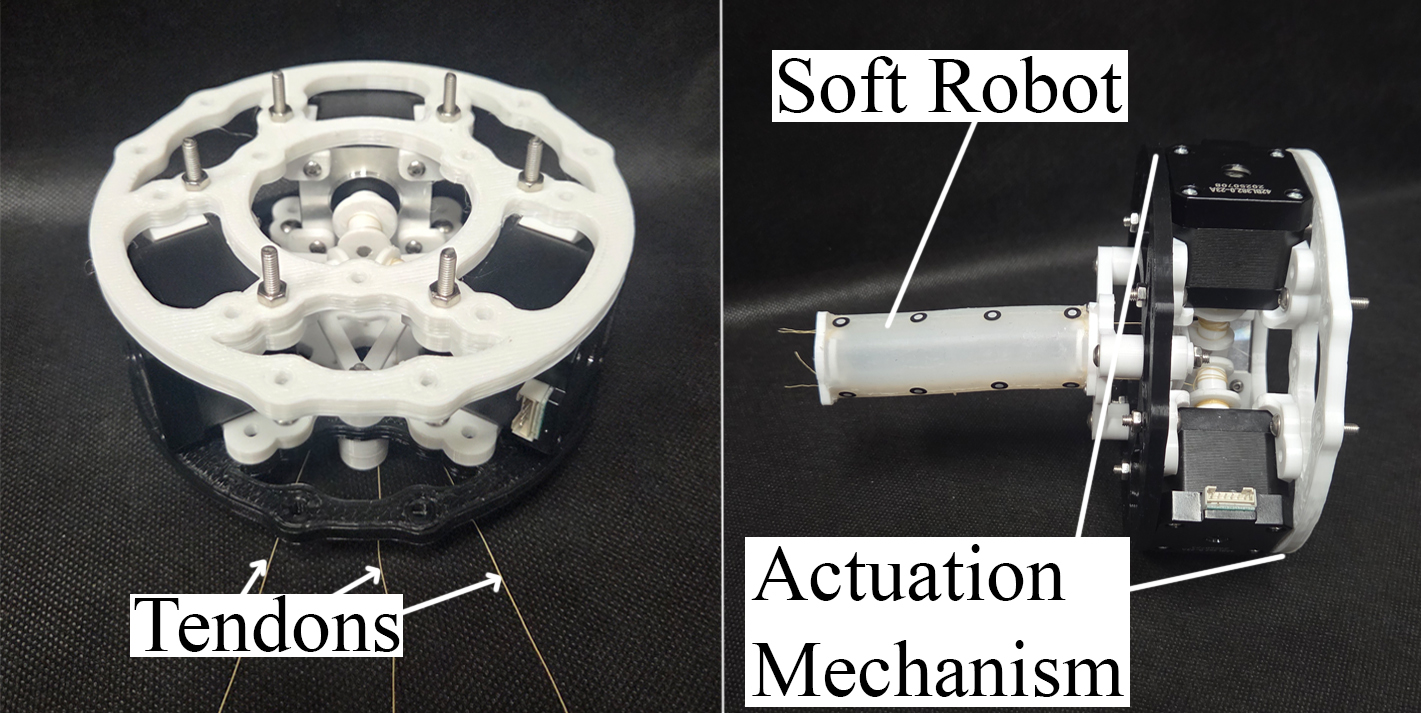}
        \caption{Single-segment assembly with tendons exiting the motor module.}
        \label{fig:actuation_module_b}
    \end{subfigure}
    \hfill
    \begin{subfigure}{0.32\textwidth}
        \centering
        \includegraphics[width=\textwidth]{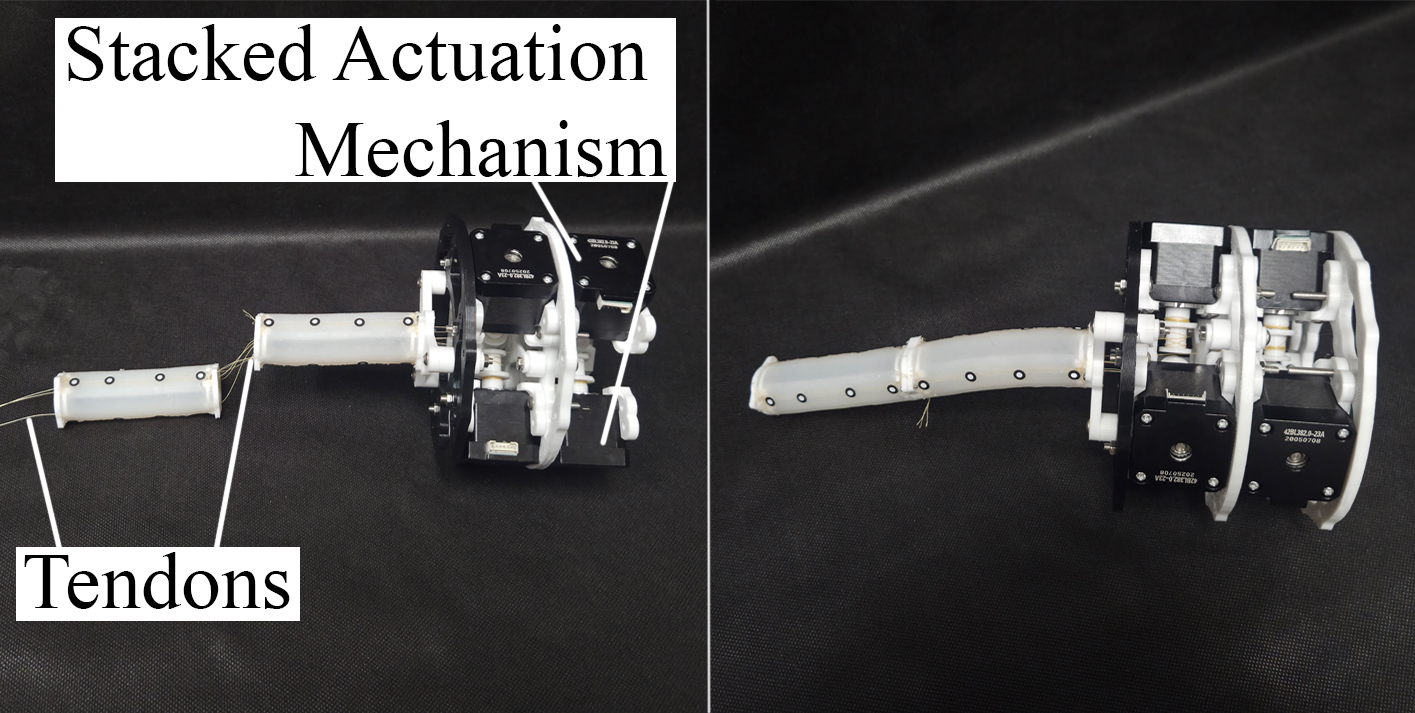}
        \caption{Two-segment configuration with added motors and routed tendons.}
        \label{fig:actuation_module_c}
    \end{subfigure}
    \hfill
    \begin{subfigure}{0.32\textwidth}
        \centering
        \includegraphics[width=\textwidth]{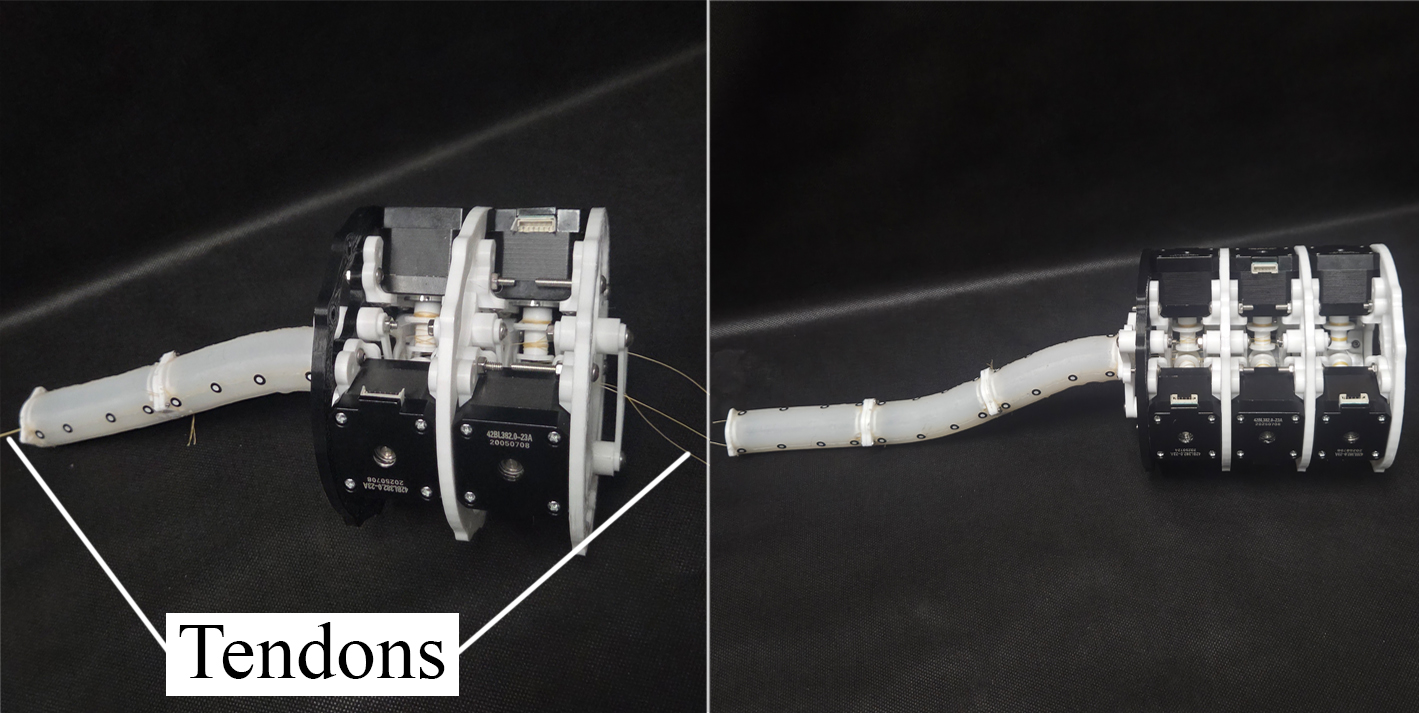}
        \caption{Three-segment configuration with extended tendons and stacked setup.}
        \label{fig:actuation_module_d}
    \end{subfigure}
     \caption{Modular design and assembly of the soft robotic arm. (a) Exploded and assembled views of the actuation module. (b)–(d) Progressive assembly of one-, two-, and three-segment configurations.}
    \label{fig:actuation_module}
\end{figure*}
% \vspace{-10pt}

\subsection{Multi-segment Configuration}
The primary design objective of the proposed system is \textit{modular scalability}. Each soft segment is stackable, allowing identical modules to be connected in series to extend reach. Each segment is casted using Ecoflex 00-50, which provides an effective balance between stiffness for force transmission and compliance. Each segment is independently actuated by three tendons driven by stepper motors mounted on a 3D-printed frame. The motors are secured using custom 3D-printed brackets, and the soft arm is attached to the frame using nuts and bolts. This combination of three motors and one soft segment constitutes a single actuation level, as illustrated in Fig.~\ref{fig:actuation_module_b}.
To extend the arm, three new tendons are passed through the backbone of the existing segment. One end of each tendon is connected to a set of stepper motors, while the other end is routed through the tendon channels of the newly attached segment. The two segments are joined using three M2 screws inserted through the end-cap holes. The new motor brackets can be mounted on top of the previous stage by loosening, aligning, and retightening the existing hardware. The two- and three-segment assemblies are shown in Figs.~\ref{fig:actuation_module_c} and~\ref{fig:actuation_module_d}, respectively.

\subsection{Control Electronics}
\begin{figure} [H]
    \centering
    \includegraphics[width=0.9\linewidth]{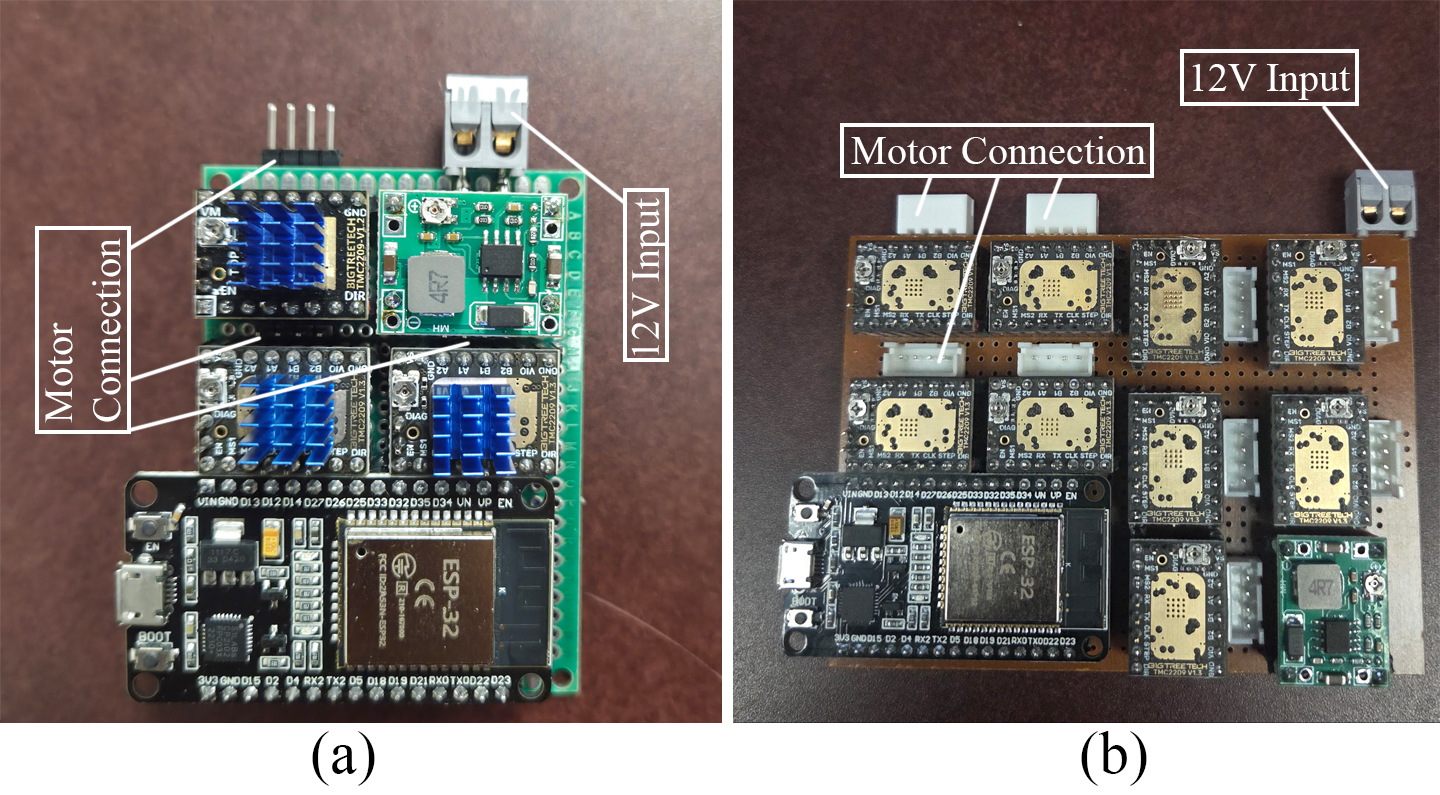}
    \caption{Controller of the modular soft robotic arm: (a) control board for a single-segment system and (b) complete setup for three-segment actuation.}
    \label{fig:control_electronics}
\end{figure}

As shown in Fig.~\ref{fig:control_electronics}, the control board consists of an ESP32 development board featuring a dual-core Tensilica Xtensa LX6 processor and multiple GPIO interfaces, that serves as the primary unit, receiving actuation commands from a laptop for coordinated operation of the soft robotic arm \cite{Yang2019}.

\vspace{-7pt}
\vspace{-14pt}

Each segment is actuated by three NEMA~17 bipolar stepper motors. Each motor is rated at 45~N$\cdot$cm torque. The motors are driven by TMC2209 stepper motor drivers. The ESP32 and motor drivers are mounted on a veroboard with appropriate wiring connections, together with an efficient buck converter that provides a regulated power supply to all modules. 
Each control board (Fig.~\ref{fig:control_electronics}(a)) functions as a self-contained unit capable of independently driving one arm segment. Boards can be daisy-chained via two-wire I\textsuperscript{2}C for scalable multi-segment control. A nine-motor-driver circuit board (Fig.~\ref{fig:control_electronics}(b)) was used for multi-segment testing.

% \vspace{-40pt}
\section{EXPERIMENTAL SETUP}
% \vspace{-10pt}
\subsection{Actuation and Control}
% A microcontroller serves as the primary control unit, receiving actuation commands from a laptop for coordinated operation of the soft robot \cite{Yang2019}. Three NEMA~17 bipolar stepper motors (45~N$\cdot$cm torque) are used to actuate each segment of the soft robotic arm. TMC2209 motor drivers are used to drive each of the stepper motors, while an ESP32 (30-pin DevKit) serves as the control unit. The DevKit and the TMC2209 drivers are mounted on a veroboard with proper wiring connections, along with a DC–DC~360 converter to provide a regulated power supply to the modules, as shown in Fig.~\ref{fig:control_electronics}.

During the payload testing, a laptop was used to send commands to the ESP32 to actuate a single tendon upto a specific distance. For the workspace and maximum radius evaluation experiments, the system was operated in an open-loop configuration, where each actuation command directly pulls a specific tendon to produce bending motion. No inverse kinematics or curvature estimation is implemented; instead, all motion is empirically controlled through visual observation and motion-capture feedback.
%This study therefore focuses on the experimental evaluation of workspace expansion and material performance of the proposed modular system, rather than on analytical modeling

\subsection{Workspace}
\begin{figure}[H]
    \centering
    \includegraphics[width=0.8\linewidth]{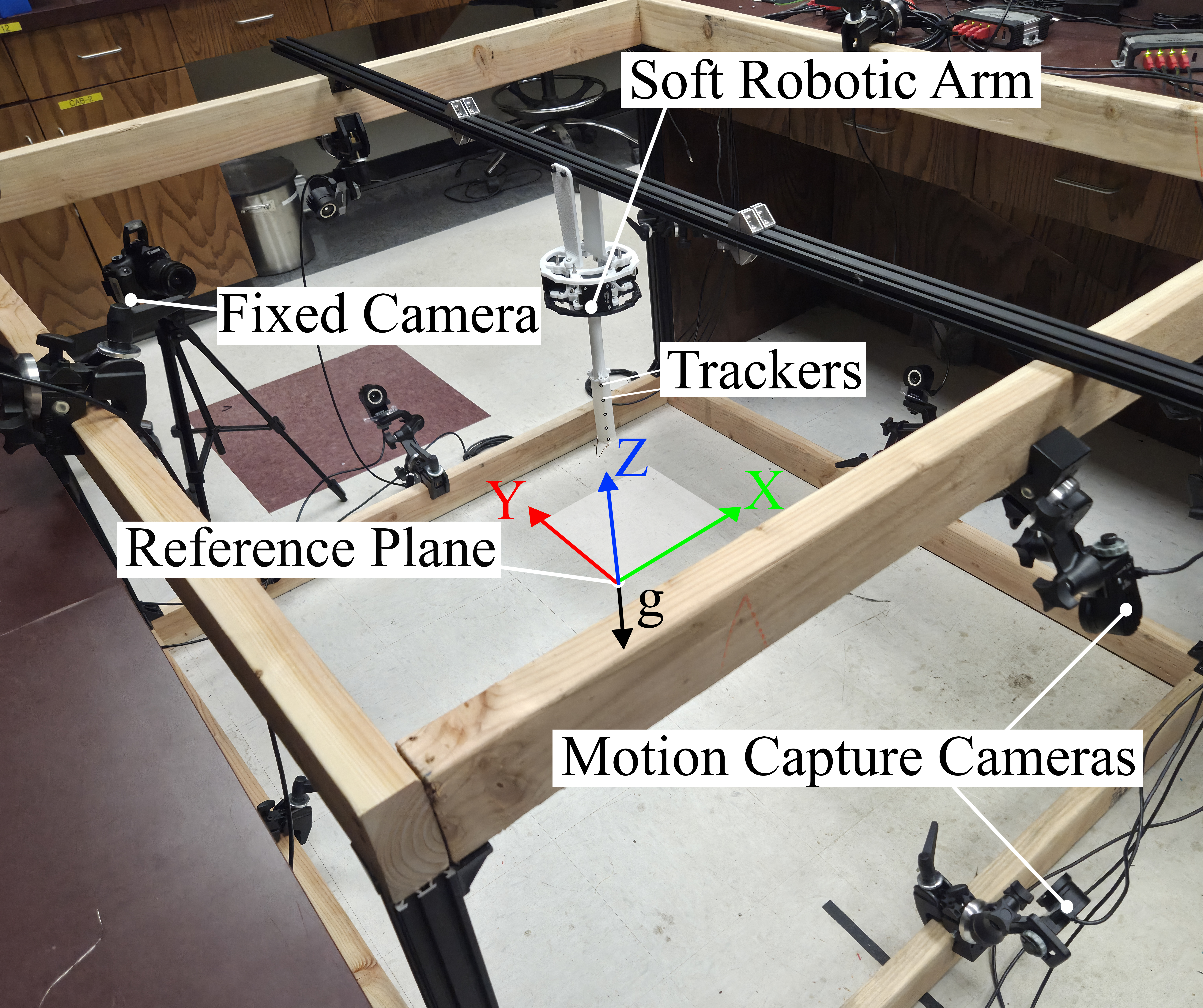}
    \caption{Experimental setup with soft robotic arm, tracking markers, motion capture cameras, and RGB camera. }
    \label{fig:workspace_setup}
\end{figure}

The three-dimensional workspace of the soft robotic arm was experimentally characterized by applying open-loop actuation commands. The arm was mounted at the center of a wooden frame surrounded by twelve OptiTrack FLEX13 motion-capture cameras, with an additional fixed RGB camera positioned to capture visual footage of the motion. The overall configuration of the experimental setup is shown in Fig.~\ref{fig:workspace_setup}, where the X–Y plane defines the reference plane and the Z-axis represents the vertical direction opposite to the direction of gravity(g). Motion-capture cameras were arranged to maintain continuous visibility of the arm during deformation. A 3D-printed extension was inserted between the soft arm and the motor frame to prevent occlusion of reflective markers.  
Experiments were conducted with single-, two-, and three-segment configurations to examine how modular stacking affects the overall reachable workspace. The recorded motion data were converted into three-dimensional point clouds, illustrating the progressive expansion of the workspace as additional segments were integrated.
%
% Data were acquired using OptiTrack Motive and subsequently processed in MATLAB for post-analysis.
% By providing open-loop commands to the soft robot, its three-dimensional workspace can be generated experimentally \cite{Wei2024}. The soft robotic arm is mounted in the center of a wooden frame surrounded by twelve OptiTrack FLEX13 motion-capture cameras, with an additional fixed RGB camera positioned to record visual footage of the motion. The complete workspace arrangement is shown in Fig.~\ref{fig:workspace_setup}.

% \vspace{-5pt}
%\vspace{-5pt}

% Motion-capture cameras were positioned and spaced to ensure continuous visibility of the soft robotic arm during motion. An 3D printed extension part was placed between the soft arm and the motor frame so that the motor frame does not occlude the reflective markers on the soft robotic arm. Each segment of the arm is equipped with twelve reflective markers, each 3 mm in diameter, to ensure accurate position tracking. Using MATLAB together with OptiTrack Motive, the movement data was collected for post-processing. Experiments were conducted for a single-segment arm, a two-segment arm, a three-segment soft robotic arm, and three-segment dual soft robotic arms. The collected data were processed in MATLAB to generate a three-dimensional point cloud, illustrating the expansion of workspace with the addition of multiple-segments and multi-arm.
\vspace{-7pt}
\begin{figure*}[!t]
    \centering
    \includegraphics[width=0.95\textwidth]{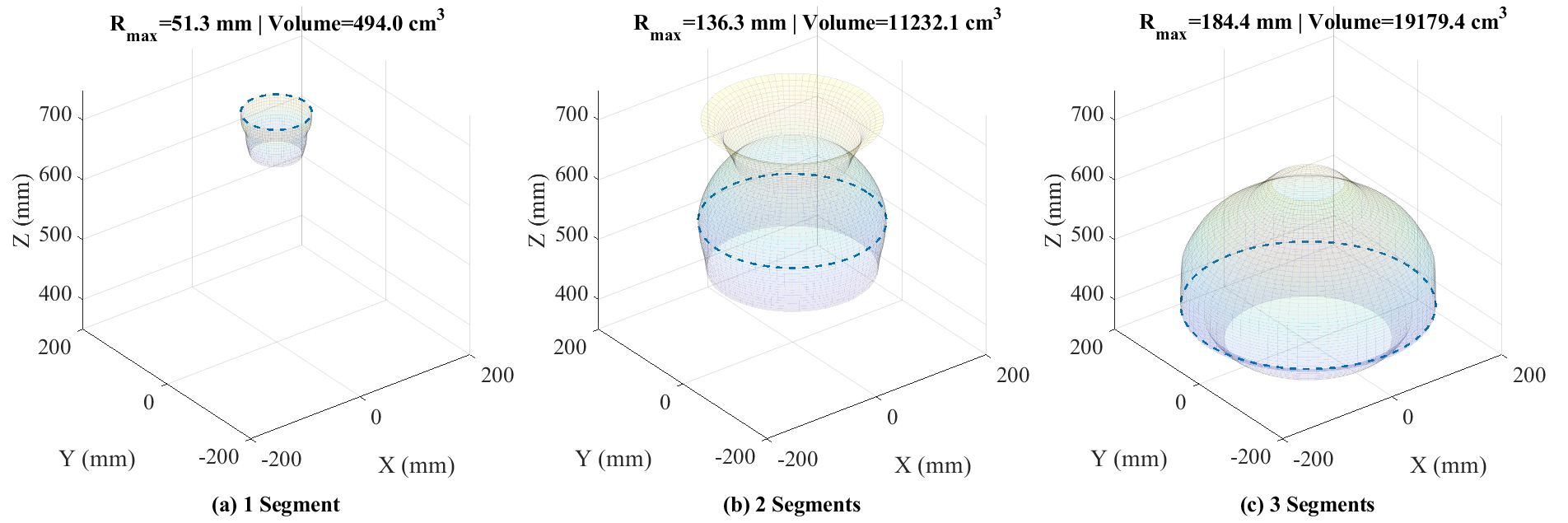}
    \caption{Workspace evaluation of the modular soft robotic arm showing (a) single-, (b) two-, and (c) three-segment configurations. Each subplot presents the workspace envelope and corresponding maximum radial reach ($R_{\max}$).}
    \label{fig:workspace_scaling}
\end{figure*}
\vspace{-14pt}

\subsection{Payload Testing Setup}

To quantify how elastomer stiffness regulates deformation under external loading and therefore achievable payload, the second design objective was to evaluate the payload capacity of the soft robotic arm as a function of material stiffness. To achieve this, a 120 mm-long segment was fabricated to improve the visibility and accuracy of bending and deflection measurements under load during the experiments. Three variants were fabricated using Ecoflex 00-10, 00-30, and 00-50, spanning a range from low to high stiffness. By maintaining identical geometry while varying only the silicone density, the influence of material stiffness and density on payload capacity, tip deflection, and overall rigidity could be directly quantified. During testing, 10~g, 20~g, 50~g, 100~g, and 200~g were applied at the tip of each segment while a single tendon was actuated by 45 mm. Each arm segment was instrumented with 3-mm reflective markers, with five markers placed at the tip to enable accurate motion tracking and curvature estimation. The motion capture setup measured the resulting bending angle and vertical tip displacement, while a force gauge recorded tendon tension, enabling evaluation of the payload-dependent mechanical performance. This comparative analysis between different silicone stiffness provides insights into the trade-off between compliance and load-bearing capability, guiding material selection for modular soft robotic applications.
% The second design objective was to evaluate payload capacity as a function of material softness. For this purpose, a 120 mm-long segment was fabricated to enhance the visibility and measurability of bending under load. Three versions were produced using Ecoflex 00-10, 00-30, and 00-50, covering a range from very soft to structurally supportive. By maintaining identical geometry and varying only the silicone type, the influence of material stiffness and density on payload capacity, tip deflection, and overall usable rigidity was directly assessed.

%To evaluate payload capacity, extended single-segment soft robotic arms fabricated from silicones with three different stiffness levels were tested. During payload experiments, the tendon was manually pulled by 45~mm, marked directly on the tendon. To reduce variability associated with manual actuation, five consecutive trials were conducted for each condition. Tip-mounted payloads of 10~g, 20~g, 50~g, 100~g, and 200~g were applied. Each arm segment was instrumented with 3-mm reflective markers, with five markers placed at the tip to enable accurate motion tracking and curvature estimation. Tendon tension was measured using a digital force gauge, while bending angle and vertical displacement were computed directly from motion-capture data. Experiments were performed using Ecoflex~00-10, 00-30, and 00-50 to assess the influence of material stiffness on actuation behavior and load-bearing performance.
%
For multi-segment loading capacity experiment, 100~mm-long segments were tested without external payloads. Instead, one and then two identical segments were sequentially attached to the distal end of the first segment, increasing the total number of segments from one to two and then to three. The tendon was pulled by 40~mm for five consecutive trials. Reflective markers remained attached to the tip of the base segment to measure deformation and vertical displacement resulting from increased structural self-weight.

% For payload evaluation, longer single-segment soft robotic arms fabricated with three different silicone densities were tested. Each arm segment was fitted with fifteen reflective markers, which were tracked by the same motion-capture setup.

% For this, a single-segment soft robotic arm with a small copper wire hook attached at its tip is mounted to the wooden frame. Different weights, ranging from 10~g to 200~g, were suspended from the hook. The arm was actuated in a single direction using open-loop control while the deformation and tip motion were recorded. A secondary camera was placed perpendicular to the plane of motion to capture the deflection behavior during each test.

% The actuation characteristics of the soft robotic arm were examined using three different silicone materials: Ecoflex 00-10, Ecoflex 00-30, and Ecoflex 00-50. Each actuator was driven under identical current conditions to isolate the effect of material stiffness. The motor current was limited to 1~A using the potentiometer on the TMC2209 driver, with each stepper motor drawing around 0.4~A during normal operation and up to 0.5~A when stalled. This ensured that the applied torque and actuation energy remained consistent across all material samples.

\section{Results \& Discussion}
To assess the performance of the proposed modular soft robotic arm, two primary characterizations were conducted:
(1) workspace scaling with increasing segment count, and
(2) actuation behavior under different silicone stiffness levels.

\subsection{Workspace characterization of modular configurations}
%The effect of modular stacking on the workspace was evaluated using motion-capture data. Tests were performed for one-, two-, and three-segment configurations, and the recorded marker trajectories were analyzed to quantify the reachable workspace. During each experiment, the distal end was actuated through its full bending range while the base remained fixed. The filtered tip trajectories were used to estimate the maximum radial reach ($R_{\max}$), planar workspace area, and three-dimensional workspace volume. 

The experimental setup and workspace envelopes for all configurations are shown in Fig.~\ref{fig:workspace_scaling}. In each experiment, the distal end was actuated through its full bending range while the base remained fixed. Since the system operates in open-loop, the workspace was 
swept by sequentially actuating each tendon to its maximum displacement, guiding the 
tip to its furthest reachable position from the arm's central axis as observed visually 
and confirmed via motion-capture feedback. The filtered tip trajectories were used to estimate the maximum radial reach ($R_{\max}$), from which the circular planar workspace area ($\pi R_{\max}^2$) and 3D workspace volume were derived by integrating circular cross-sections along the arm's axial direction. The single segment achieved a maximum radial reach of 51.3~mm and a planar workspace area of approximately 8.26$\times$10\textsuperscript{3}~mm\textsuperscript{2}. The recorded trajectories revealed a near-circular bending path, consistent with the constant-curvature assumption in continuum kinematics \cite{GarrigaCasanovas2018}. The motion was smooth and repeatable, confirming structural stability and precise tendon control. Adding a second segment increased the reach to 136.3~mm and expanded the planar workspace area to 5.84$\times$10\textsuperscript{4}~mm\textsuperscript{2}, representing a 7.1-fold gain. The three-segment configuration achieved a planar area of 1.07$\times$10\textsuperscript{5}~mm\textsuperscript{2} and a workspace volume of 1.92$\times$10\textsuperscript{7}~mm\textsuperscript{3}, indicating that workspace scalability arises primarily from enhanced spatial flexibility rather than simple radial extension. The summarized results are presented in Table~\ref{tab:workspace_data}.

% The single-segment arm achieved a maximum radial reach of 48.3~mm and a planar workspace area of approximately 7.32$\times$10\textsuperscript{3}~mm\textsuperscript{2}. When a second segment was added, the reach increased to 122.3~mm and the workspace expanded to 8.32$\times$10\textsuperscript{4}~mm\textsuperscript{2}, representing an 11.4-fold increase in reachable area compared to the single-segment configuration. The three-segment arm exhibited a maximum reach of 106.5~mm but achieved the largest overall workspace, with an estimated planar area of 1.33$\times$10\textsuperscript{5}~mm\textsuperscript{2} and a revolved volume of approximately 6.7$\times$10\textsuperscript{6}~mm\textsuperscript{3}. These findings indicate that the workspace scalability arises primarily from enhanced 3D flexibility and spatial coverage rather than pure radial extension. The summarized metrics for all configurations are presented in Table~\ref{tab:workspace_data}.

\begin{table}[H]
\centering
\caption{Workspace evaluation of single- and multi-segment configurations.}
\label{tab:workspace_data}
\large
\resizebox{\linewidth}{!}{
\begin{tabular}{lccc}
\toprule
\textbf{Configuration} & \textbf{$R_{\max}$ (mm)} & \textbf{Planar Area (mm$^2$)} & \textbf{Workspace Volume (mm$^3$)} \\
\midrule
One segment              & 51.3  & 8.26$\times$10\textsuperscript{3}  & 4.94$\times$10\textsuperscript{5} \\
Two segments             & 136.3 & 5.84$\times$10\textsuperscript{4}  & 1.12$\times$10\textsuperscript{7} \\
Three segments             & 184.4 & 1.07$\times$10\textsuperscript{5}  & 1.92$\times$10\textsuperscript{7} \\
\bottomrule
\end{tabular}}
\end{table}

\subsection{Effect of Silicone Stiffness  on Actuation Behavior}

\begin{figure}[H]
\centering
\includegraphics[width=0.9\linewidth]{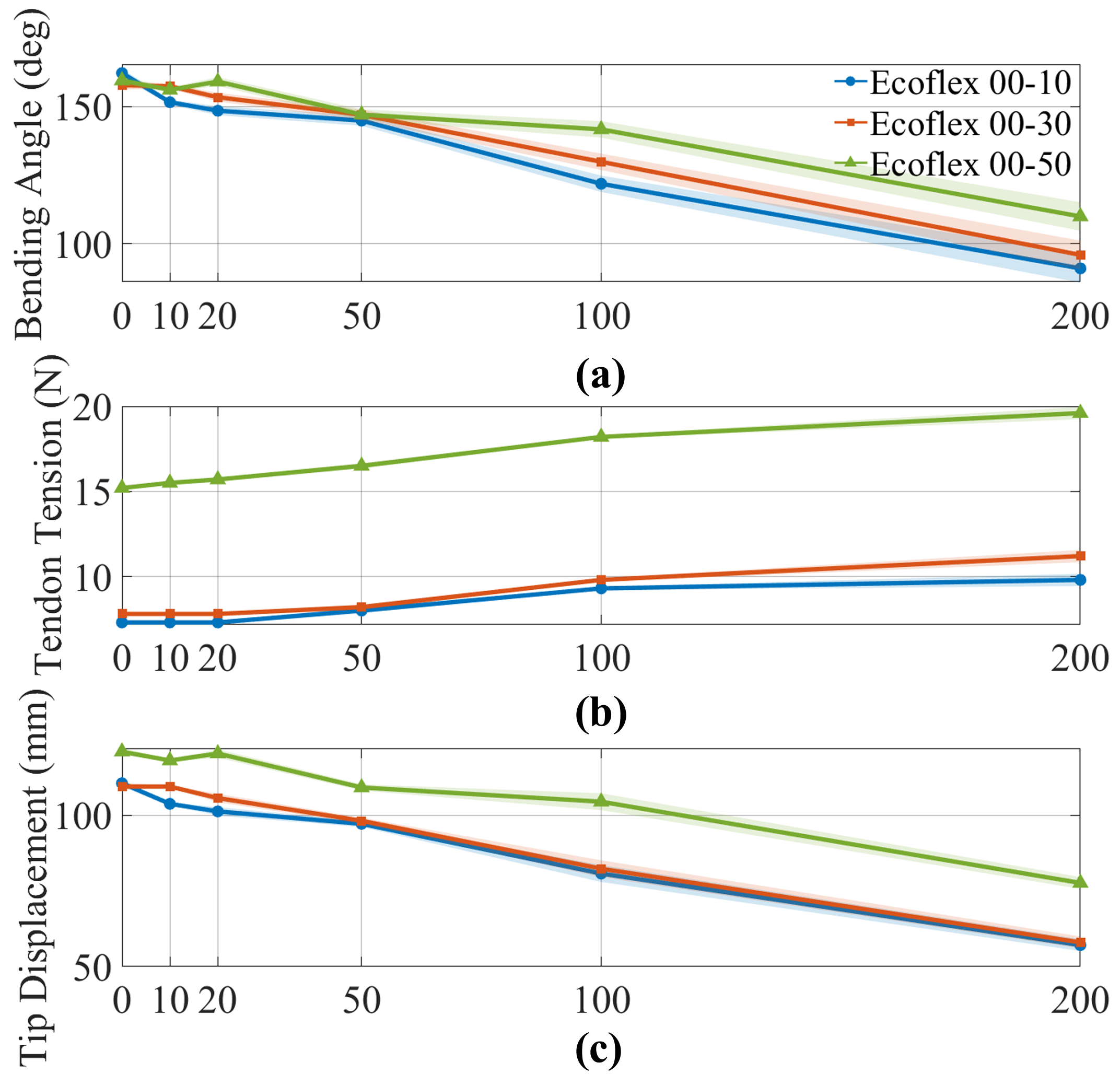}
\caption{Effect of payload on (a) bending angle, (b) tendon tension, and (c) vertical displacement  across silicone stiffness levels.}
\label{fig:summary}
\end{figure}

Figure~\ref{fig:summary} summarizes the single-segment payload experiments under a fixed tendon displacement of 45~mm for the three silicone formulations. The shaded regions represent ±1 standard deviation across five repeated measurements for each condition, while the markers denote the corresponding mean values, indicating that the observed trends are consistent despite relatively small experimental variability. As payload increased from 0 to 200~g, all actuators exhibited a consistent reduction in bending angle and vertical tip displacement, accompanied by an increase in required tendon tension. For example, the Ecoflex~00-10 actuator showed a drop in bending angle from approximately 162° to 91° and a decrease in vertical displacement from about 110~mm to 57~mm, while the required tendon tension increased from about 7.3~N to 9.8~N. In contrast, the stiffer Ecoflex~00-50 maintained a much higher structural stiffness, requiring substantially larger tendon forces (15–20~N) but preserving greater geometric stability under load.

% \vspace{-10pt}
% \vspace{-14pt} 

\begin{figure*}[t]
\centering
\includegraphics[width=0.95\textwidth]{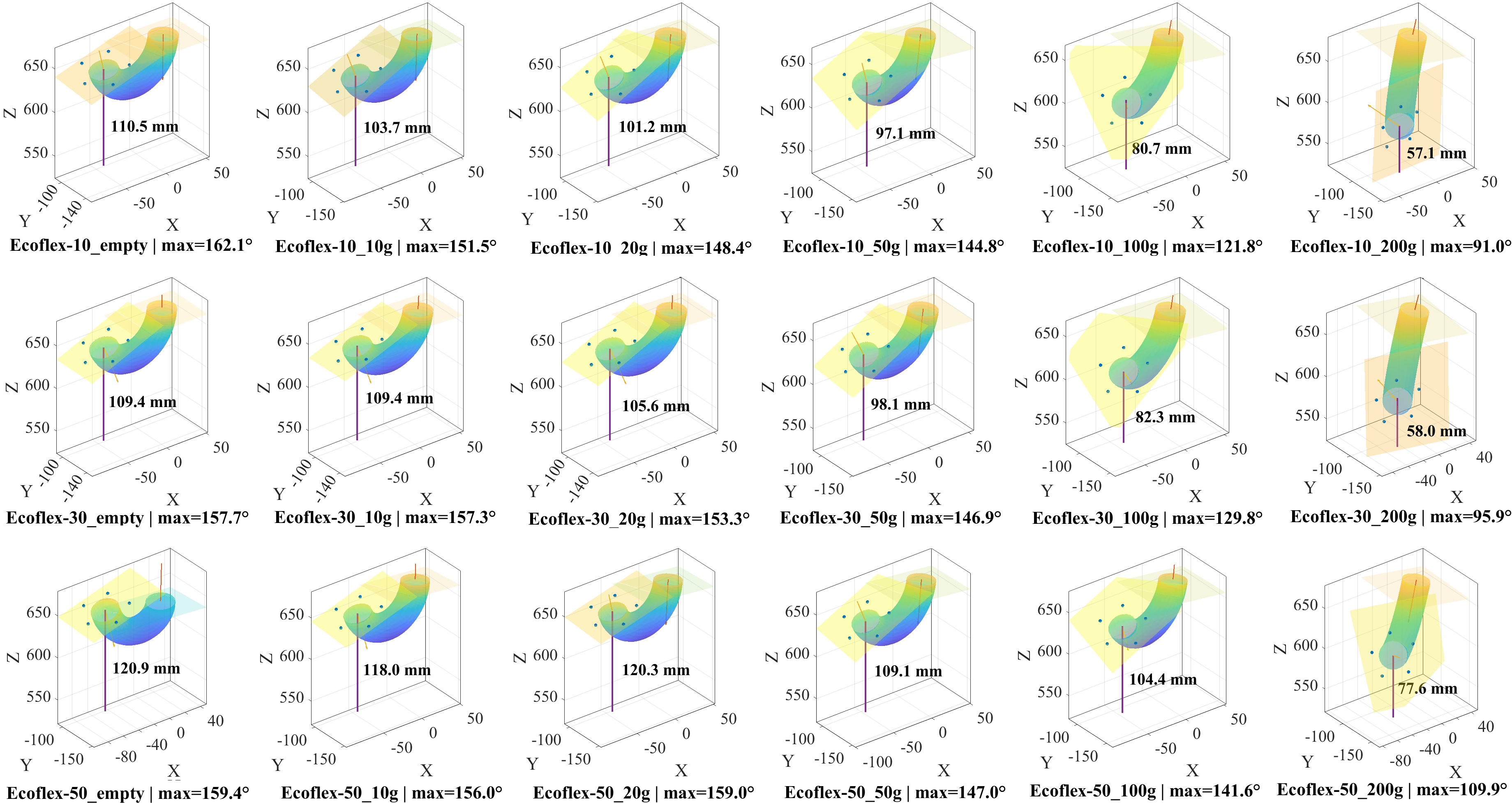}
\caption{Bending curvature and deformation of the modular soft robotic arm under varying payloads for different silicone stiffness levels. While Ecoflex-00-10 and Ecoflex-00-30 achieve higher bending angles, Ecoflex-00-50 exhibits greater linear displacement along the Z-axis, highlighting a clear trade-off between angular flexibility and vertical load-bearing deformation.}
\label{fig:full15}
\end{figure*} 
The detailed payload-dependent trends for bending angle, tendon tension, and vertical displacement are shown in Fig.~\ref{fig:full15}. Under increasing payload, the softer materials (Ecoflex~00-10 and 00-30) exhibited pronounced backbone sagging and non-uniform deformation, while the stiffer Ecoflex~00-50 preserved its overall shape but at the cost of reduced bending amplitude. Although bending angles were computed assuming constant curvature from the five tip markers, higher loads frequently caused deviations from this idealized shape, indicating that external loading can violate the constant-curvature assumption commonly used in continuum kinematic models.

\begin{figure}[H]
\centering
\includegraphics[width=0.85\linewidth]{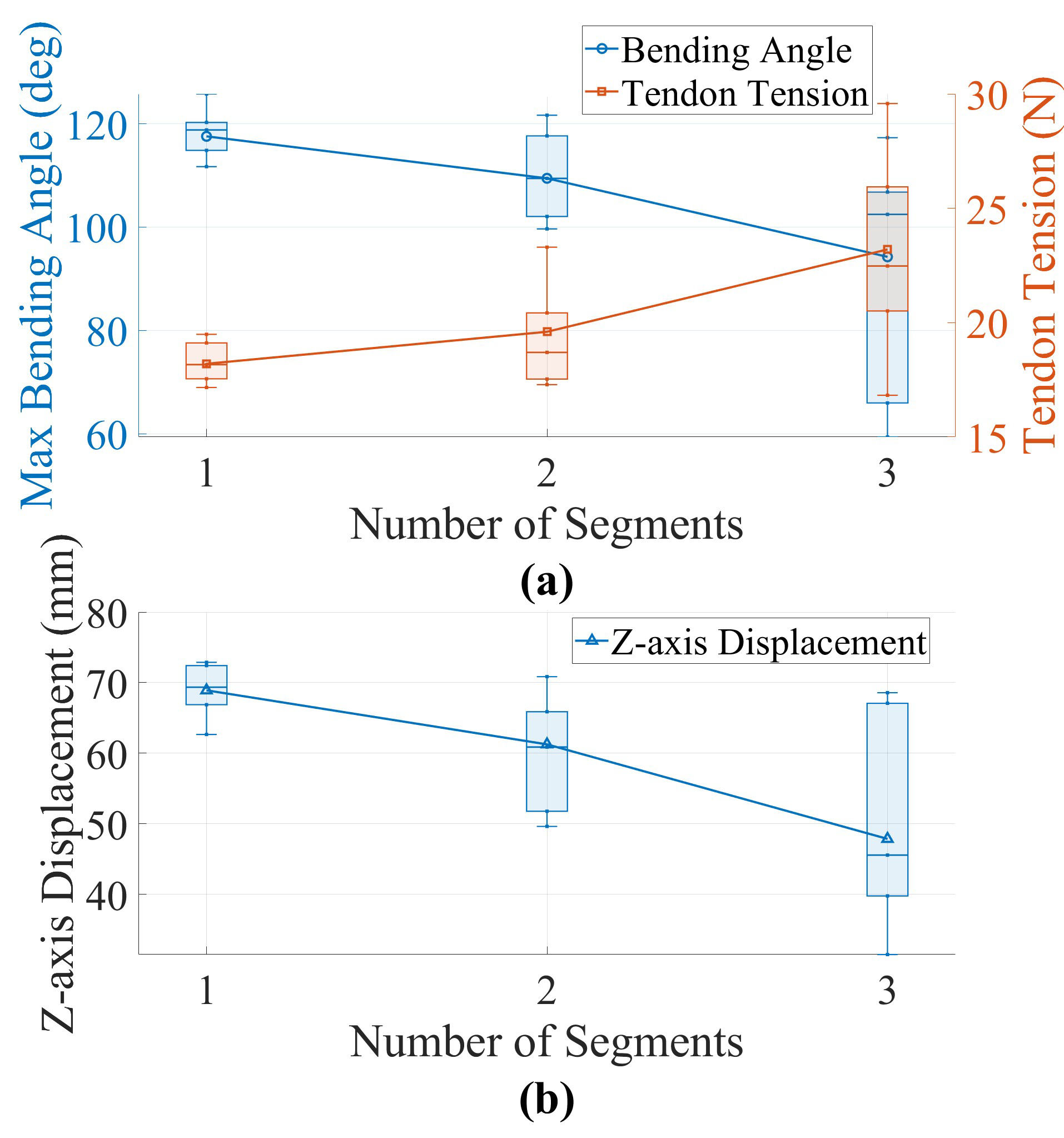}
\caption{Effect of segment stacking on (a) bending angle and tendon tension and (b) vertical (Z-axis) displacement.}
\label{fig:multiseg}
\end{figure}

The effect of structural loading becomes more evident in the multi-segment experiments (Fig.~\ref{fig:multiseg}). As identical segments were stacked from one to two and then three, the effective self-weight and compliance of the system increased, resulting in reduced bending, higher tendon tension, and decreased vertical displacement at the base segment. When normalized, both the bending angle and vertical displacement exhibit nearly identical degradation trends with increasing segment count, indicating a strongly coupled geometric scaling effect. In contrast, tendon tension follows a different and more nonlinear growth pattern, suggesting that actuation effort scales more aggressively than kinematic loss as structural load accumulates. This demonstrates that payload and self-weight accumulate rapidly in modular configurations, imposing a practical limit on scalable segment stacking.

%\subsection{Design implications for modular soft manipulators}
%The results highlight key considerations for the design of modular, cable-driven soft manipulators. Modular stacking substantially expands the reachable workspace, with gains primarily arising from increased spatial flexibility rather than radial extension alone. Material stiffness strongly influences actuation behavior: softer silicones enable larger bending but exhibit reduced geometric stability under load, whereas stiffer silicones improve payload capacity at the expense of bending range. In multi-segment configurations, accumulated self-weight and compliance degrade bending performance and increase actuation effort nonlinearly, imposing practical limits on scalable segment stacking. These findings underscore the need to jointly consider material stiffness and modular configuration when designing scalable soft robotic arms.

\vspace{-10pt}
% \vspace{-14pt}

% As shown in Fig.~\ref{fig:full15}, the bending curvature decreased noticeably with increasing silicone stiffness. The actuator molded with Ecoflex~10 exhibited the largest bending deformation and responded readily to actuation even under small loads. Ecoflex 00-30 produced a moderate bend, while Ecoflex 00-50 was considerably stiffer, showing only slight deflection even with a 10~g load. However, under higher payloads (100~g and 200~g), the lower-density samples (Ecoflex 00-10 and 00-30) exhibited structural deformation along the backbone, losing their original shape during actuation. This behavior indicates that the softer materials were unable to maintain internal structural stability when loaded, causing the tendons to bear most of the mechanical stress.

% In contrast, the Ecoflex 00-50 actuator maintained its geometric stability and resisted backbone deformation even at 200~g, despite exhibiting limited bending. This demonstrates that higher-density silicones provide greater load-bearing capability and shape retention, while softer materials favor flexibility but suffer from reduced structural stiffness under external loads. Overall, the results reveal a clear trade-off between compliance and load support: Ecoflex 00-10 and 00-30 offer greater actuation range but poorer structural integrity under weight, whereas Ecoflex 00-50 prioritizes stiffness and form retention at the cost of bending amplitude.

\section{CONCLUSIONS}
This work presented a modular, cable-driven soft robotic arm with a reconfigurable multi-segment architecture that enabled scalable extension without disassembly. Experimental results showed that softer Ecoflex~00-10 and 00-30 achieved larger bending but exhibited backbone sagging and kinematic degradation under heavier load, whereas stiffer Ecoflex~00-50 improved load-bearing capacity and geometric stability at the expense of bending range and higher pulling force requirement. Although segment stacking increased reach, accumulated self-weight and compliance increased actuation effort and reduced bending performance, imposing practical limits on scalable extension. These observations provided quantitative guidance for material selection and modular configuration in multi-segment soft manipulators. These results highlight key considerations for the design of modular, cable-driven soft manipulators, and together provide quantitative guidance for material selection and modular configuration in multi-segment systems. In future work, the theoretical kinematics of tendon-driven continuum manipulators will be incorporated using the constant-curvature assumption \cite{Webster2010, Walker2005} to enable more precise control and trajectory planning based on the experimentally characterized material and geometric properties of the proposed modular platform.

% This work introduced a novel modular cable-driven soft robotic arm that enables multi-segment reconfigurability—a capability rarely achieved in tendon-actuated soft robotic systems. The modular design allows new segments to be integrated seamlessly without disassembling existing components. Experimental evaluation demonstrated that increasing the number of connected segments significantly expanded the arm’s workspace and bending range, validating the scalability of the approach. Material characterization revealed that softer silicones such as Ecoflex 00-10 and 00-30 enhance bending flexibility, whereas the stiffer Ecoflex 00-50 provides improved structural rigidity and load-bearing stability, underscoring the trade-off between compliance and stiffness in soft actuator performance.

% Future work will focus on optimizing the backbone geometry and wall thickness to strengthen the structure while retaining the flexibility benefits of softer materials. In parallel, efforts will be directed toward developing a closed-loop control framework integrating sensor feedback and model-based actuation, enabling more precise and coordinated motion across multi-segment and multi-arm configurations.

\vspace{-5pt}
\bibliography{ref}
\bibliographystyle{IEEEtran}

\end{document}